\documentclass[preprint,12pt]{elsarticle}
\usepackage[hidelinks]{hyperref}
\usepackage{xcolor}
\usepackage{subcaption}
\usepackage{adjustbox}
\newcommand{\modelname}{Structural Restricted Boltzmann Machine }
\newcommand{\modelacronym}{SBM }
\newcommand{\modelacronymp}{SBM}

\usepackage{amssymb}
\journal{Neurocomputing}

\begin{document}

\begin{frontmatter}



\title{\modelname for image denoising and classification}


\author[inst1]{Arkaitz Bidaurrazaga}

\affiliation[inst1]{BCAM - Basque Center for Applied Mathematics,
Mazarredo, 14 E48009 Bilbao, Basque Country – Spain.
E-mail: abidaurrazaga@bcamath.org}

\author[inst1]{Aritz Pérez}
\author[inst2]{Roberto Santana}

\affiliation[inst2]{organization={Intelligent Systems Group},%Department and Organization
            addressline={Paseo de Manuel  Lardizabal, 1 }, 
            city={Donostia},
            postcode={20018}, 
            state={Gipuzkoa},
            country={Spain}}

\begin{abstract}
Restricted Boltzmann Machines are generative models that consist of a layer of hidden variables connected to another layer of visible units, and they are used to model the distribution over visible variables. In order to gain a higher representability power, many hidden units are commonly used, which, in combination with a large number of visible units, leads to a high number of trainable parameters. In this work we introduce the \modelname model, which taking advantage of the structure of the data in hand, constrains connections of hidden units to subsets of visible units in order to reduce significantly the number of trainable parameters, without compromising performance. As a possible area of application, we focus on image modelling. Based on the nature of the images, the structure of the connections is given in terms of spatial neighbourhoods over the pixels of the image that constitute the visible variables of the model. We conduct extensive experiments on various image domains. Image denoising is evaluated with corrupted images from the MNIST dataset. The generative power of our models is compared to vanilla RBMs, as well as their classification performance, which is assessed with five different image domains. Results show that our proposed model has a faster and more stable training, while also obtaining better results compared to an RBM with no constrained connections between its visible and hidden units. 
\end{abstract}

\iffalse
\begin{graphicalabstract}
\includegraphics{grabs}
\end{graphicalabstract}

\begin{highlights}
\item The \textit{\modelname}(\modelacronymp) model is formalised, generalising previous works on sparse RBM and opening a new framework where the structure is determined for hidden-visible connections.
    \item We focus on \modelacronymp s for images, and therefore describe sensible choices of subsets of visible units as pixel neighbourhoods, these subsets may not be disjoint, as will be shown in the formalisation.
    \item We address standard tasks such as density estimation, image denoising and image classification. 
    \item Results show that \modelacronymp s obtain better results compared to RBM in certain domains, \modelacronymp s having much fewer trainable parameters compared to the RBMs. This fact highlights the advantage of \modelacronymp s in terms of performance and computational complexity.
\end{highlights}
\fi


 \end{frontmatter}

\section{Introduction}

Many Machine Learning tasks address the problem of approximating the underlying distribution of a dataset, where said distribution may be used to generate new samples or to classify new instances. Among the existing models which perform these tasks, the \textit{Restricted Boltzmann Machine} (RBM) is one of the simplest models with competitive results \cite{gehler2006rate,hinton2007recognize}. Yet RBMs may require a high number of parameters to represent the probability distribution, which entails a computationally expensive and slow training. In this work we propose an RBM model with sparse weight matrix, thus reducing the number of trainable parameters. A sparse weight matrix is achieved by constraining connections of each hidden unit to a local neighbourhood of visible units given in terms of subsets of visible units, instead of being connected to every visible unit. These connections can be restrained using prior knowledge about the problem at hand, or they could be driven from the data itself. As a possible problem domain, we work with images. Thus, we focus on using prior knowledge about images, and train our models to reconstruct and classify images. Our main goal is to show that better or at least comparable results can be achieved by sensibly pruning some hidden-visible connections from RBMs. The main contributions of this paper are: 
\begin{itemize}
    \item The \textit{\modelname}(\modelacronymp) model is formalised, generalising previous works towards sparse RBM and opening a new framework where the structure of hidden-visible connections is determined via the prior knowledge available of the problem at hand.
    \item We focus on \modelacronymp s for images, and therefore describe sensible choices of subsets of visible units as pixel neighbourhoods, these subsets may not be disjoint, as will be shown in the formalisation.
    \item We address standard tasks such as density estimation, image denoising and image classification. 
    \item Results show that \modelacronymp s obtain better results compared to RBM in certain domains, \modelacronymp s having much fewer trainable parameters compared to the RBMs. This fact highlights the advantage of \modelacronymp s in terms of performance and computational complexity.

\end{itemize}

The paper is structured as follows. The background on RBM and Related work is reviewed in this section. Section \ref{Proposed model} introduces the proposed model, expanding on how it is trained and how it is defined for images. In Section \ref{Experiments}, the datasets and models used in the experiments and the general pipeline are described. The obtained results are discussed in Section \ref{Discussion}. Finally, conclusions are given and main future goals are presented in Section \ref{Conclusion}.

\section{Preliminaries}\label{Background}
\textit{Energy-based models} (EBM) emerged in the 1980s, starting with the \textit{Hopfield network} \cite{hopfield1982neural}, which described the activation of visible binary units in a deterministic manner and served as associative memories. The \textit{Boltzmann Machine} (BM) was proposed in \cite{hinton1983optimal} as the stochastic version of the Hopfield network, and was subsequently extended to use hidden variables in order to gain representational power. Nevertheless, training and sampling from a BM with intra-layer connections is demanding, therefore restricting connections between hidden-hidden and visible-visible units was proposed in \cite{smolensky1986information}, hence the model was named Restricted Boltzmann machine. 

The \textit{Restricted Boltzmann Machine} is a probabilistic EBM with a two layer architecture in which $n_v$ visible stochastic binary units $\textbf{v}\in \{0,1\}^{n_v}$ are connected to $n_h$ hidden stochastic binary units $\textbf{h}\in\{0,1\}^{n_h}$. Even though RBMs were initially used for binary visible variables, they have been generalised to non-binary and continuous variables \cite{welling2004exponential,courville2011spike}. The energy function of an RBM (binary units) is defined as

\begin{eqnarray}\label{RBM energy}
    E(\textbf{v},\textbf{h};\theta)=-\sum_{i,j}v_iW_{ij}h_j-\sum_{i}v_ia_i-\sum_{j}h_jb_j,
\end{eqnarray}
which is parametrised by the set of parameters $\theta = \{W,\textbf{a},\textbf{b}\}$, and the sum indices $i,j$ run from 1 to $n_v,n_h$, respectively. Given the energy function, an RBM models the probability distribution following the Boltzmann distribution

\begin{eqnarray}
    q(\textbf{v},\textbf{h};\theta)=\frac{\exp(-E(\textbf{v},\textbf{h};\theta))}{Z(\theta)},
\end{eqnarray}
where $\theta$ are the parameters of the model, and $Z(\theta)=\sum_{\textbf{v}'}\sum_{\textbf{h}'}\exp(-E(\textbf{v}',\textbf{h}';\theta))$ is the normalisation constant known as the partition function, which in practice, even for moderate values of $n_v$ and $n_h$, is intractable. 

An important property of RBM with binary hidden units is that marginalisation over hidden vectors can be performed analytically. This allows us to describe the probability distribution over the visible units in terms of the free energy $F(\textbf{v};\theta)$, as in Eq. (\ref{Prob Free energy}) and (\ref{Free energy}).

\begin{eqnarray}
    q(\textbf{v};\theta)&=&\frac{\exp(-F(\textbf{v};\theta))}{Z(\theta)}\ ,\ Z(\theta) = \sum_{\textbf{v}'}\exp(-F(\textbf{v}';\theta)), \label{Prob Free energy}\\
    F(\textbf{v};\theta) &=& -\sum_i v_ia_i -\sum_j \ln (1+\exp(\sum_i v_iW_{ij}+b_j)), \label{Free energy}
\end{eqnarray}
where in Eq. (\ref{Prob Free energy}) the partition function $Z$ is computed by adding for every visible configuration.

A common objective when training an RBM is to adjust the parameters $\theta$ such that the model can approximate the empirical probability distribution of the data $\hat{p}(\textbf{v})$. There are several methods to train an RBM \cite{montufar2016restricted,upadhya2019overview}, but a popular one is by optimising the parameters to minimise the Kullback-Leibler (KL) divergence with respect to the empirical distribution. In this manner, the formalisation of the objective when training RBMs is written as follows:

\begin{eqnarray}
    \theta^* &=& \arg\min_{\theta} D_{KL}(q(\textbf{v};\theta)||\hat{p}(\textbf{v}))
\end{eqnarray}
where $D_{KL}(q(\textbf{v},\theta)||\hat{p}(\textbf{v}))=\mathbb{E}_{q}[\log (q(\textbf{v},\theta)/\hat{p}(\textbf{v}))]$ is the KL divergence between the distributions $q$ and $\hat{p}$.

In \cite{upadhya2019overview}, authors demonstrate that minimising the KL-divergence is equivalent to maximising the \textit{loglikelihood} (LL) of the training set. \textit{Stochastic mini-batch Gradient Descent} (SGD) is commonly used to maximise LL (actually to minimise $-\mathcal{L}(\theta)=\mathbb{E}_{\hat{p}}[-\log q(\textbf{v};\theta)]$), where the gradients used to update each parameter are given by Eqs. \ref{W gradient}-\ref{Hidden gradient}.

\begin{eqnarray}\label{Param Gradients}
    \frac{\partial\mathcal{L}(\theta)}{\partial W_{ij}}&=&\mathbb{E}_{\hat{p}}[v_ih_j]-\mathbb{E}_{q}[v_ih_j],\label{W gradient}\\
    \frac{\partial\mathcal{L}(\theta)}{\partial a_{i}}&=&\mathbb{E}_{\hat{p}}[v_i]-\mathbb{E}_{q}[v_i],\label{Visible gradient}\\
    \frac{\partial\mathcal{L}(\theta)}{\partial b_{j}}&=&\mathbb{E}_{\hat{p}}[h_j]-\mathbb{E}_{q}[h_j], \label{Hidden gradient}
\end{eqnarray}
where $\mathbb{E}_{p}$ denotes expectation over the distribution $p$, and $\hat{p}$ is the empirical distribution and $q$ is the model distribution.

\textit{Contrastive Divergence} (CD) is a common procedure proposed by Hinton \cite{hinton2002training} where, discarding a negligible term, the LL gradient is crudely approximated. CD estimates the expectation $\mathbb{E}_{q}$ via a Markov Chain Monte Carlo (MCMC) method called \textit{Gibbs sampling} (GS) \cite{geman1984stochastic}. This algorithm draws a sequence of samples that approximately follow the desired distribution when direct sampling is difficult. Because sampling hidden and visible units simultaneously from an RBM is infeasible, due to the intractable computation of the partition function, samples of $\textbf{v}$ are drawn by alternating samples between $p(\textbf{v}|\textbf{h})$ and $p(\textbf{h}|\textbf{v})$, where the chain is initialised with a data point $\textbf{v}^{(n)}$. Moreover, each hidden/visible unit can be sampled in parallel, because every unit is conditionally independent from each other given a visible/hidden state. This property arises from the fact that in RBMs intra-layer connections are forbidden. Given a hidden/visible state, the probability of a visible/hidden state to be equal to 1 is

\begin{eqnarray}\label{RBM Probs}
    p(v_i=1|\mathbf{h})&=&\frac{p(v_i=1,\mathbf{h})}{p(v_i=0,\mathbf{h})+p(v_i=1,\mathbf{h})}\nonumber\\
    &=&\sigma(\sum_{j}W_{ij}h_j+a_i),\\
    p(h_j=1|\mathbf{v})&=&\sigma(\sum_{i}W_{ij}v_i+b_j),
\end{eqnarray}
where $\sigma(x)=(1+e^{-x})^{-1}$ is the sigmoid function. Each GS step is performed $K$ times, since the mixing time of the underlying Markov Chain could require more than a single loop in order to obtain good samples, thus the algorithm is called CD-$K$. The number of GS steps is considered as a hyperparameter, but in practice a small $K$ is used because as learning progresses the mixing time of the Markov chain decreases rapidly. Even when
using a single step, Carreira and Hinton \cite{carreira2005contrastive} have shown that CD produces a small
bias for a large speed-up in training time.

The number of hidden units $n_h$ is another hyperparameter of the RBMs as well, although some recent works \cite{loukas2019self} have proposed methods for RBMs to automatically learn the number of hidden units. Given an input of dimension $n_v$, setting $n_h$ determines the number of parameters of an RBM, which is of the order $\mathcal{O}(n_vn_h)$. Thus, for high dimensional data, the training convergence can be slow, as the computational time complexity is directly related to the number of parameters of the weight matrix $W$. In this work, we propose a sensible approach to reduce the number of trainable parameters by constraining some visible-hidden connections. 

\subsection{How to evaluate RBMs}\label{sec:evaluate RBM}

RBMs were originally proposed for generative purposes, where the probability distribution of a given dataset is approximated, from which samples could be drawn. When training each model, the mean LL of the train and validation set are evaluated. This is indeed the objective function the model optimises using mini-batch SGD with CD. For a set of visible data $\mathcal{D}$, the LL is computed as follows

\begin{eqnarray}
    \mathcal{L}(\mathcal{D};\theta)&=&\frac{1}{|\mathcal{D}|}\log \prod_{\textbf{v}\in\mathcal{D}}q(\textbf{v};\theta)=\frac{1}{|\mathcal{D}|}\log \prod_{\textbf{v}\in\mathcal{D}}\frac{\exp(-F(\textbf{v};\theta))}{Z(\theta)}\nonumber\\ 
    &=&-\frac{1}{|\mathcal{D}|}\sum_{\textbf{v}\in\mathcal{D}}F(\textbf{v};\theta)-\log Z(\theta)
\end{eqnarray}
where the \textit{Annealed Importance Sampling} (AIS) method is used to approximate the $\log Z(\theta)$ term. We follow the setting presented in \cite{salakhutdinov2008quantitative} to estimate the partition function of the RBMs with 1000 AIS iterations, although using 5 times fewer intermediate distributions. 

One of the problems these models can handle is image denoising, consequently, we assess the performance of our models on this task as well. Given an image that is the distorted version of an original image, the denoising problem consists of recovering the original image by removing noise. RBMs and \modelacronymp s trained to approximate the underlying distribution of a set of images can be used for this purpose, by means of Gibbs sampling. Images are reconstructed after a single Gibbs sampling step by setting the corrupted image as initial state, the resulting visible state is taken as the reconstructed image. The performance is measured by computing the mean square error (MSE) between the reconstructed image and the original non-corrupted image.

\subsection{Classification RBM}\label{Classification RBM}

Even though RBMs were originally introduced as generative models, they can be used for classification. RBMs are commonly trained to extract features, which are then fed to a classifier network \cite{gehler2006rate}, this can be seen as a feature extraction method. Nonetheless, RBMs can also be used as stand-alone discriminative models, as proposed by Larochelle and Bengio in \cite{larochelle2008classification}. In \cite{bi2019early}, authors use a convolutional deep BM to classify Electroencephalograms for early Alzheimer’s disease diagnosis, and in \cite{he2013facial} a two-layer deep BM which classifies facial expressions from thermal infrared images is proposed.

In this work, the Classification RBM proposed by Larochelee and Bengio in \cite{larochelle2008classification} is used to perform image classification purposes. In this model, classes are represented with a one-hot encoded vector $\textbf{y}=(1_{y=k})_{k=1}^C$, and they are added to the RBM as additional visible units. These variables are connected to the hidden units via the matrix $U$ and have biases $\textbf{c}$. Consequently, similarly to Eq. (\ref{RBM energy}), the energy of this model is 

\begin{eqnarray}\label{ClassRBM energy}
    E(\textbf{v},\textbf{y},\textbf{h};\theta)&=&-\sum_{i,j} v_iW_{ij}h_j-\sum_iv_ia_i-\sum_jh_jb_j\nonumber\\
    &&-\sum_ky_kc_k-\sum_{k,j}y_kU_{kj}h_j.
\end{eqnarray}
Which leads to the modification of the expressions of some conditional probabilities for GS

\begin{eqnarray}
    p(y_k=1|\textbf{h})&=&\frac{\exp(c_k+\sum_jU_{kj}h_j)}{\sum_{k'}\exp(c_{k'}+\sum_jU_{k'j}h_j)}, \\
    p(h_j=1|y_k,\textbf{v})&=&\sigma(\sum_{i}W_{ij}v_i+b_j+U_{kj}).
\end{eqnarray}
Deriving how to update the new parameters using CD is straightforward. However, an alternative method more suited for classification was suggested in \cite{larochelle2008classification}.

The probability of each class $y_k$ given an instance $\textbf{v}$ can be computed via

\begin{eqnarray}
    p(y_k=1|\textbf{v}) &=& \frac{\exp(c_k)\prod_{j=1}^{n_h}(1+\exp(\sum_{i}W_{ij}v_i+b_j+U_{kj}))}{\sum_{k'}\exp(c_{k'})\prod_{j=1}^{n_h}(1+\exp(\sum_{i}W_{ij}v_i+b_j+U_{k'j}))}. 
\end{eqnarray}

Authors in \cite{larochelle2008classification} proposed to train the model such that the probability of the true class given an image is maximised. Hence, the discriminative objective function can be written as in Eq. (\ref{Discriminative objective}). 
\begin{eqnarray}\label{Discriminative objective}
    \theta^* &=& \arg\min_{\theta} -\log p(y|\textbf{v};\theta)
\end{eqnarray}

The gradients of this objective function with respect to the parameters can be computed exactly as given in  Eq. (\ref{Classification gradients})

\begin{eqnarray}\label{Classification gradients}
    \frac{\partial p(y_k|\textbf{v};\theta)}{\partial \theta} &=& \sum_j \sigma(o_{kj}(\textbf{v}))\frac{\partial o_{kj}(\textbf{v})}{\partial \theta}\nonumber\\
    &-&\sum_{j,m}\sigma(o_{mj}(\textbf{v}))p(y_m|\textbf{v})\frac{\partial o_{mj}(\textbf{v})}{\partial \theta},
\end{eqnarray}
where $o_{kj}(\textbf{v})=b_j+U_{kj}+\sum_i W_{ij}v_i$. Therefore, when training our models for the classification problem, Eq. (\ref{Classification gradients}) is used when updating the parameters with SGD.

\subsection{How to evaluate classification RBMs}\label{Evaluate Classification}

Standard metrics to evaluate a classifier are used to assess classification RBMs as well. Classification accuracy is one of the most popular metrics, which evaluates the ratio between the number of well-classified instances and the total number of instances classified. This metric has the risk of not being representative of the classification performance when classes are unbalanced. Therefore, when classes are unbalanced, the balanced version of accuracy can be used, where the contribution of each class is weighted with the inverse of its frequency of appearance.

Nonetheless, since our models are able to return the probability of each class given an image, other metrics that penalise giving lower probabilities to the correct class can be used. One of these metrics is the Log-loss function
\begin{eqnarray}
    L_{log}(Y,P) = -\frac{1}{N}\sum_{i=1}^{N}\sum_{k=1}^{K}y_{i,k}\log p_{i,k},
\end{eqnarray}
where $Y$ is a vector of $N$ one-hot represented $K$ classes, so $y_{i,k}=1$ if the instance $i$ corresponds to class $k$ and 0 otherwise, and $P$ is the vector of probabilities the model assigns to each instance. The closer the Log-loss value is to 0, the more accurate and confident the predictions of the model are. In this case, the same strategy used with the accuracy is used when classes are unbalanced, assigning weights to each class according to the inverse of its frequency of appearance.

\section{Related work}\label{Related work}

In this work, we focus on exploiting the structure of the visible-hidden weight parameters, as a way of reducing the computational cost of the model while keeping a comparable performance. There are various previous works that limit the number of interactions between visible and hidden units on RBMs. For instance, using RBMs to simulate quantum many-body systems, authors in \cite{pilati2020simulating} connect each hidden unit to 2 or 3 visible units only, whereas the approach proposed in \cite{sehayek2019learnability} pretrains the model to omit small learned parameters and later fine-tune the remaining weights. Both works report an improvement on the results over the baseline RBM. In \cite{mittelman2014structured}, Mittelman et al. train a Recurrent Temporal RBM, and learn a structure which determines whether the interactions between visible and hidden units are masked. They show that this method significantly improves the prediction performance when the number of visible units is large and the training set is small. 

In \cite{ji2014enhancing,cho2012tikhonov}, authors added different regularisation terms to the objective function, these techniques aimed at achieving sparser hidden activations. That is, reducing the number of active hidden units when being sampled from a visible state, which enhanced the discriminative performance when feeding the hidden representations to a classifier. Using sparse connections has already been proposed by Mocanu et al. in \cite{mocanu2016topological}, where a random bipartite graph with the scale-free property is constructed to build sparse connections between hidden and visible units. More recently, they have proposed the Sparse Evolutionary Training RBM in \cite{mocanu2018scalable}, where sparse connections adapt during training. In this work, we focus on structures that do not evolve as the training continues, but which are fixed from the beginning. Moreover, we will assess our models with images, where our model constrains hidden connections to certain pixel neighbourhoods, which considerably reduces the number of trainable parameters while maintaining the dependencies between pixels within their respective neighbourhood. A previous work exists where authors mention the idea of using neighbourhoods of pixels connected to each hidden variable when training a Hidden Markov Random Field (HMRF) \cite{kunsch1995hidden}, with the task of generating images of two different textures, however, the authors do not develop this idea any further. 

Although RBMs and MLPs are very different paradigms, in MLPs various methods that try to reduce the number of parameters of the networks have been proposed as well. Some of these methods similarities with the strategies to enforce sparsity in RBMs, and therefore we briefly review them here. Authors in \cite{frankle2019lottery} suggest a similar approach to \cite{sehayek2019learnability} based on what they call the \textit{The Lottery Ticket Hypothesis}, while they retrain the parameters by applying a mask instead of fine-tuning them. 

In contrast with \cite{sehayek2019learnability}, authors in \cite{lee2019snip} propose the \textit{Single-shot Network Pruning} method, which does not require the model to be pretrained in order to determine which connection should be pruned. They compute the gradient of the loss function with respect to some auxiliary variables that define if a connection is \textit{on} or \textit{off}, and then the importance of each connection is decided by the magnitude of its respective gradient. In \cite{bengio2000taking}, Bengio et al. excluded some connections from a MLP via statistical tests, and showed that significant improvements can be obtained in terms of out-of-sample \textit{loglikelihood} by pruning the network. In \cite{jain2020locally}, authors generate images with a convolutional network, and apply different masks to the convolutional weights depending on the order in which the pixels are generated, obtaining globally coherent image completions.

\section{\modelname}\label{Proposed model}
In order to control the number of trainable parameters of an RBM, we propose an alternative RBM model that constrains hidden connections to certain subsets of input data variables (visible units). These connections can be restrained using prior knowledge, or can be driven from the data itself. We call this model the \modelname (\modelacronymp). 

Given a set of visible variables $\mathcal{V}$ and a hidden unit $h_j$, we define a collection of subsets of variables $\mathcal{V}(h_j)\subset\mathcal{V}$, and connect each hidden unit $h_j$ to the corresponding subset. In this way, we guarantee to obtain a sparse weight matrix $W_{ij}$, where $W_{ij}\neq 0$ only if $h_j$ is connected to the visible unit $v_i\in\mathcal{V}(h_j)$. Accordingly, subsets of hidden units $\mathcal{H}(v_i)\subset\mathcal{H}$ which connect to the corresponding visible unit $v_i$ can be defined. In this manner, the hidden-visible interaction term in Eq. (\ref{RBM energy}) can be rewritten as follows:

\begin{eqnarray}
    \sum_{j}\sum_{v_i\in \mathcal{V}(h_j)} v_i W_{ij} h_j = \sum_{i} \sum_{h_j\in \mathcal{H}(v_i)} v_i W_{ij} h_j.
\end{eqnarray}
In general, these subsets will be defined in the visible units $\mathcal{V}(h_j)$, where structural knowledge from the data can be applied. The number of trainable parameters for an RBM is $\mathcal{O}(n_vn_h)$, whereas considering we build equally sized subsets, i.e., $\forall j:|\mathcal{V}(h_j)|= V \ll n_v$, the number of parameters of a \modelacronym is $\mathcal{O}(V\cdot n_h)$. Our main objective is to build sensible choices of small neighbourhoods, which define the subsets $\mathcal{V}(h_j)$, such that the number of trainable parameters is reduced considerably, while maintaining or improving performance on the problem the model attempts to solve.

\subsection{\modelacronym for images}
The introduced \modelacronym models take advantage of the underlying structure of the problem to reduce the number of trainable parameters. As an area of application, we focus on images in this work. We assume that in the image problems addressed there is a spatial relationship between the value of a pixel and its neighbourhood. Hence, each hidden unit may focus on different patches of visible units (pixels), and thus, each patch/neighbourhood of visible units will take the role of $\mathcal{V}(h_j)$. In order to define these sets, given a pixel $\textbf{u}=(u_x,u_y)\in \mathcal{I}$ within a $d\times d$ image, we use the \textit{Chebyshev distance} between two pixels $d(\textbf{u},\textbf{u}')=\max (|u_x-u'_x|,|u_y-u'_y|)$ to define the neighbourhood. The \textit{neighbourhood} of a pixel $\textbf{u}'$ is defined as $\mathcal{N}(\textbf{u}')=\{\textbf{u}|d(\textbf{u},\textbf{u}')\leq w; \textbf{u}'\in \mathcal{I}\}$, where $w$ is the \textit{window size} parameter. This parameter determines the size of the square shaped neighbourhood of visible units of side length $2w+1$ pixels, and thus the number of visible variables within the neighbourhood of $\textbf{u}$ is $|\mathcal{N}(\textbf{u})|=(2w+1)^2$. These sets of neighbour pixels take the role of the aforementioned subsets $\mathcal{V}(h_j)$.

\begin{figure}[ht]
    \centering
    \includegraphics[width = \textwidth]{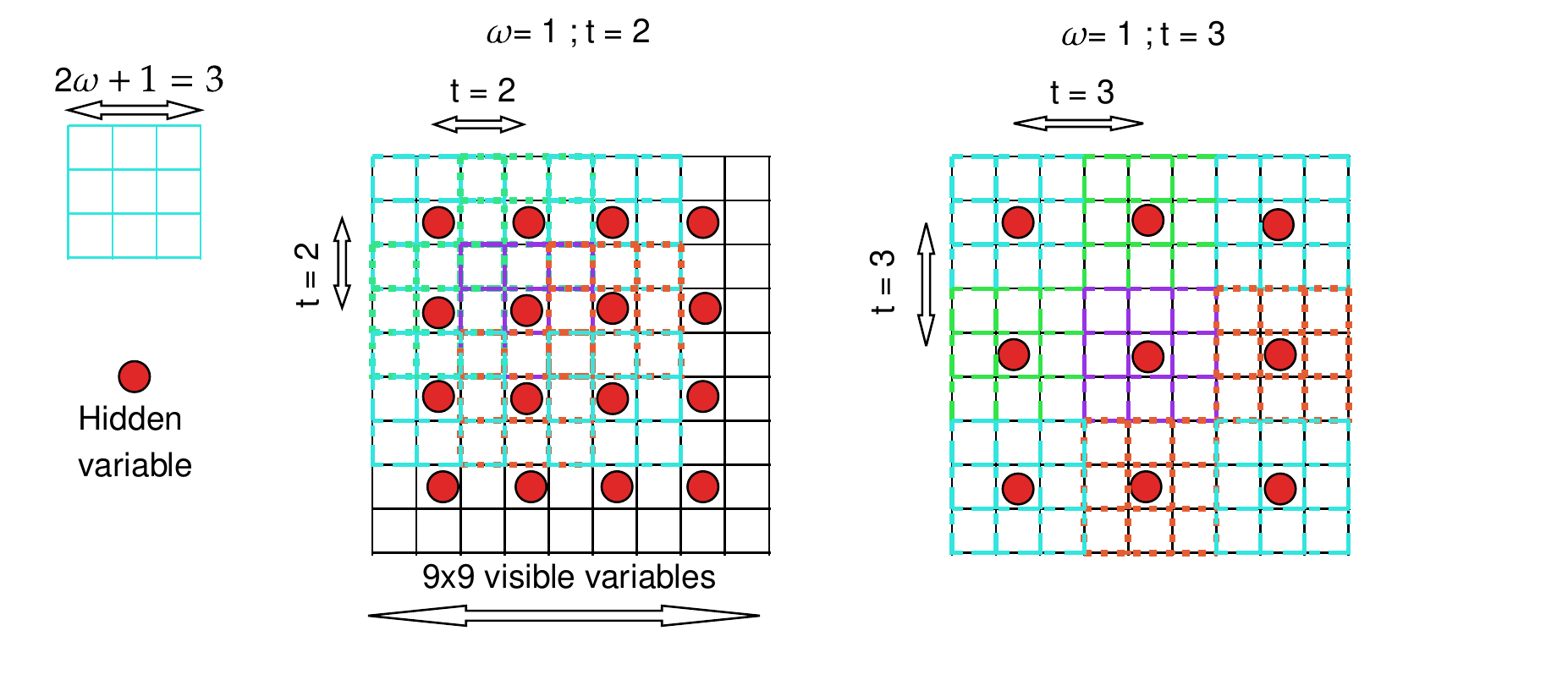}
    \caption{Illustration of some neighbourhood structures. The window size $w$ is the distance (Chebyshev) between the centre of the neighbourhood and its farthest pixel. The stride $t$ determines the distance between each neighbourhood centre, where a hidden unit is located to represent that it is connected only to those pixels within the neighbourhood. }
    \label{fig:stride and window}
\end{figure}

However, in order to further control the number of hidden units, we introduce another parameter called \textit{stride} parameter $t$, which determines how distant each patch of visible units is from each other. The stride parameter, in combination with the window parameter, allows us to control the number of hidden units. These neighbourhoods are distributed such that the centres of each neighbourhood are separated by a minimum distance of $t$. Thus, the set of \textit{neighbourhood centres}, which is a function of the window and stride parameters $(w,t)$, is defined as $\mathcal{C}(w,t)=\{\textbf{u}|w\leq u_x,u_y\leq d-w \wedge d(\textbf{u},\textbf{u}')\geq t; \textbf{u},\textbf{u}'\in \mathcal{I}\}$. In order to clarify and remove any ambiguity, we specify that $\textbf{u}=(w,w)$ shall be an element of the set $\mathcal{C}(w,t)$. As can be seen in Figure \ref{fig:stride and window}, each hidden unit is only connected to a single subset of pixels and, for illustrative purposes, they are located at the centres given by $\mathcal{C}(w,t)$. Notice that, for our model, there is no need to determine the number of hidden units $n_h$, since it is governed by the parameters ($w,t$), where $|\mathcal{C}(w,t)|=n_h$. Therefore, we define an \modelacronym model with parameters $(w,t)$ as $\mathcal{M}(w,t)$, where each hidden unit $h_j$ is connected only to the visible units $v_i\in\mathcal{N}(\textbf{u})$, and there is a one-to-one correspondence between every hidden unit and every $\textbf{u}\in \mathcal{C}(w,t)$.

Nonetheless, there is no need to be restrained to a single set of hidden units determined by $(w,t)$. Several sets of hidden units connected to visible neighbourhoods using different parameters can be stacked, e.g., it is possible to combine two sets of constrained hidden units, where we represent the new stacked model as $\mathcal{M}(w_1,t_1;w_2,t_2)=\mathcal{M}(w_1,t_1)\cup \mathcal{M}(w_2,t_2)$. 

Such a definition of neighbourhoods may resemble filters from a Convolutional Neural Network (CNN), and although the proposed procedure is inspired by convolutional filters it has nothing to do with them. CNN filters are applied sequentially to patches of pixels in order to obtain a different shaped output, while our neighbourhoods determine connections to a hidden unit (or a range of hidden units) which does not entail a sequential application of a filter.

\section{Experiments}\label{Experiments}

In this section we specify the datasets, the structure of the RBM and \modelacronym models, and the learning procedure used to perform the experiments presented in Section \ref{Discussion}.

\subsection{Datasets}\label{sec:Datasets}
 
We compare the performance of our models using benchmark datasets MNIST, FashionMNIST, OrganAMNIST, OrganCMNIST and OrganSMNIST. All of them consist of $28\times 28$ images in grey scale, and even though pixels take discrete values between 0 and 255, they can be easily normalised to the range $[0,1]$, so we can treat the intensities as probabilities. This strategy was used for digit recognition in \cite{mayraz2000recognizing}. Characteristics of each dataset are shown in Table \ref{tab:Datasets}.

MNIST consists of images of handwritten digits. It is a benchmark dataset used for the image classification task, and for unsupervised learning with generative models as well. Moreover, in \cite{mu2019mnist}, authors contribute by distorting the original MNIST handwritten digits with different corruption types, such that these new images could be used to assess the robustness of classification models against different kinds of noises. This dataset is called the MNIST-C. We use these corrupted images to analyse how well our models perform on the denoising labour as well as on classification. Even though 15 corruptions are available, we use 8 in our experiments. The 7 discarded corruptions involve affine transformation of images, for instance, \textit{rotation} or \textit{scale}. It is important that corruptions do not involve the translation, rotation or scaling of the original image, because even though the reconstructed image could be a plausible image for the human eye, the denoising error we use to evaluate execution would be biased and would not be faithful. This may happen because reconstructing an image may remove noisy pixels, but probably will not rotate images to their natural orientation or scale them to the original size. On the other hand, corruptions such as \textit{stripe}, \textit{fog} and \textit{brightness} have high intensity values on the border pixels, which, we have observed, destabilise the RBMs while sampling, since they learn images with border pixels of intensity 0. Therefore, we use corruptions that simply add different types of non-affine noise to the original image or apply a blurring filter, which are the following: \textit{shot\_noise, impulse\_noise, glass\_blur, motion\_blur, spatter, dotted\_line} and \textit{zigzag}.

The FashionMNIST dataset is a set of grey scale images of 10 fashion products, e.g., shoes, shirts and bags. It is commonly used for supervised problems as the original MNIST dataset. Even though neither MNIST nor FashionMNIST have a validation set, during training 10000 instances from the training set are used as the validation set. Thus, the actual size of the train size of MNIST and FashionMNIST is 50000 if we do not state otherwise.

The last three datasets are three of the many available from the MedMNIST dataset. The OrganAMNIST is based on 3D computed tomography (CT) images from Liver Tumor Segmentation Benchmark. Authors used bounding-box annotations of 11 body organs from another study to obtain the organ labels. They also cropped 2D images from the centre slices of the 3D bounding boxes in axial views to obtain the images. Images were resized into 28×28 to perform multi-class classification of 11 body organs. OrganCMNIST and OrganSMNIST were obtained using the same procedure, where the difference comes from cropping the 3D images in Coronal and Sagittal views, respectively. It is worth mentioning that these datasets present unbalanced classes, having two predominant classes. Therefore, when assessing their classification performance, the weighted version of the Log-loss mentioned in Section \ref{Evaluate Classification} is employed.

\begin{table}[t]
\resizebox{\textwidth}{!}{
\begin{tabular}{|ccccc|}
\hline
\textbf{Dataset} & \textbf{Domain}   & \textbf{Classes} & \textbf{Train/Val/Test} & \textbf{ML Task}\\ \hline
MNIST\cite{lecun1998gradient}           & Handwritten Digits            & 10                  & 50000/10000/10000   & C/U                \\
MNIST-C\cite{mu2019mnist}           & Corrupted Handwritten Digits            & 10                  & 60000/-/10000  & C/D                   \\
FashionMNIST\cite{DBLP:journals/corr/abs-1708-07747}     & Fashion products  & 10                  & 50000/10000/10000   & C/U                \\
OrganAMNIST\cite{medmnistv1}     & Axial Liver Tumor Segmentation  & 11                  & 34581/6491/17778   & C                \\
OrganCMNIST\cite{medmnistv1}   & Coronal Liver Tumor Segmentation      & 11                   & 13000/2392/8268          & C          \\
OrganSMNIST\cite{medmnistv1}      & Sagittal Liver Tumor Segmentation & 11                   & 13940/2452/8829             & C        \\\hline
\end{tabular}
}
\caption{Description of the datasets used for the experiments. The Machine Learning task each dataset is used for is stated in the column \textit{ML Task}, where each letter stands for Classification (C), Unsupervised learning (U) and Denoising (D). The MNIST-C dataset is not used for training, only the test set is used for evaluation of models trained using the original MNIST dataset.}
\label{tab:Datasets}
\end{table}

\subsection{Model structure}

Six \modelacronym models defined with different $w,t$ parameters were compared, where, for each \modelacronymp, we train another RBM as shown in Table \ref{tab:Model params table}, using the same number of hidden units as their related \modelacronymp, we refer to them as \textit{twin RBM}. The parameters that define each model are shown in Table \ref{tab:Model params table}. The choice of the $(w,t)$ values was motivated to attain a sufficient number of connections between visible and hidden units so that representational power was not lost, while trying to reduce them as much as possible. Every \modelacronym in Table \ref{tab:Model params table} requires between 6\% and 10\% of the number of parameters compared to its twin RBM. The \modelacronym and RBM models have been implemented in Python using TensorFlow-2 \cite{tensorflow2015} and the implementation is freely available\footnote{It can be accessed at \url{https://github.com/arkano29/StructuralRBM}.}.

\begin{table}[ht]
   \centering
\begin{tabular}{|lcccc|}
\hline
\multicolumn{1}{|c}{\textbf{Model}} & \textbf{$\mathcal{M}(w,t)$}                                & \textbf{\textbf{Nº hidden}} & \textbf{\textbf{Nº parameters}} & \% \\ \hline
$RBM_{121}$                         & --                                                         & 121                                          & $9.49\cdot10^4$  &   --     \\
$SBM_{121}$                         & $\mathcal{M}(4,2)$                                         & 121                                          & $9.6\cdot10^3$  &   10      \\ \hline
$RBM_{144}$                         & --                                                         & 144                                          & $1.13\cdot10^5$  &   --   \\
$SBM_{144}$                         & $\mathcal{M}(3,2)$                                         & 144                                          & $6.89\cdot10^3$  &   6      \\ \hline
$RBM_{265}$                         & --                                                         & 265                                          & $2.08\cdot10^5$  &  --          \\
$SBM_{265}$                         & $\mathcal{M}(3,2;4,2)$                                     & 265                                          & $1.65\cdot10^4$  &  8      \\ \hline
$RBM_{441}$                         & --                                                         & 441                                          & $3.46\cdot10^5$  &    --         \\
$SBM_{441}$                         & $\mathcal{M}(4,1)$                                         & 441                                          & $3.53\cdot10^4$  &    10      \\ \hline
$RBM_{562}$                          & --                                                         & 562                                           & $4.41\cdot10^5$  &  --        \\
$SBM_{562}$                          & $\mathcal{M}(4,2;4,1)$                                         & 562                                           & $4.49\cdot10^4$  &  10   \\ \hline
$RBM_{585}$                          & --                                                         & 585                                           & $4.59\cdot10^5$   & --     \\
$SBM_{585}$                          & $\mathcal{M}(3,2;4,1)$                                         & 585                                           & $4.22\cdot10^4$  & 9    \\ \hline
\end{tabular}
\caption{Trained \modelacronym models and their respective twin RBMs. The number of hidden units and the number of trainable parameters of each model are displayed as well. The structure for sparse connections of \modelacronymp s is given in the column $\mathcal{M}(w,t)$. Recall the notation for stacked models being defined as $\mathcal{M}(w_1,t_1;w_2,t_2)=\mathcal{M}(w_1,t_1)\cup \mathcal{M}(w_2,t_2)$.}
\label{tab:Model params table}
\end{table}

\subsection{Learning procedure}\label{sec:Pipeline}

We train the models using mini-batch SGD, using CD with a single Gibbs sampling step for unsupervised learning, while parameters were updated following Eq. (\ref{Classification gradients}) on the classification problem. The introduced sparse model can be trained using the learning procedures of the fully connected RBM.

For the sake of increasing the training speed, we apply momentum \cite{qian1999momentum} to the parameter updates. At training epoch $t$, the parameters are updated according to $\Delta \theta (t)=v_i(t)=\beta v_i(t-1)-\epsilon\frac{\partial E}{\partial\theta}(t)$, where the first term is the accumulated momentum and the second term is the gradient of the objective function with respect to the parameter $\theta$, and $\epsilon$ and $\beta$ are the learning rate and the momentum hyperparameter, respectively. The learning rate is set to $\epsilon=0.1$ and momentum $\beta=0.9$.

We executed the experiments for each model with 10-repeated hold outs, using different seeds, which influences the initialisation of the weight parameters. The weight parameters are initialised randomly following the uniform distribution $U[-\sqrt{6}/\sqrt{n_v+n_h},\sqrt{6}/\sqrt{n_v+n_h}]$, as proposed in \cite{glorot2010understanding} by Glorot and Bengio, while biases are initialised to 0. As stated by Hinton \cite{hinton2012practical}, using a large batch size with mini-batch SGD is a serious mistake, thus he recommends using sizes of the order of tens. Therefore, the batch size is set to 16, and batches of data instances are drawn randomly from the train set at each update step.

Models are trained with 10000 updates of the parameters, and the parameters with the best metric on the validation set during training, which depends on the objective of the training, are chosen for model assessment on the test set. For the unsupervised learning problem, LL is used as the metric to choose parameters, while for the classification problem, the Log-loss is used.

\subsection{Results and discussion}\label{Discussion}

In this section we investigate the performance of \modelacronymp s and their twin RBMs for the tasks of probability density estimation, image denoising and image classification.

We train our models to fulfil the unsupervised task of estimating the probability density of the digit and fashion product images introduced in Sec. \ref{sec:Datasets}. This is fulfilled by training our models to approximate the underlying distribution of the dataset, where the approximation quality is measured using the LL of the model over the test set. The performance of denoising digit images was also assessed. Elapsed time when computing the CD gradient is compared between RBMs and \modelacronymp s as well. 

Finally, the models have been trained to classify images for different domains, as explained in Sec. \ref{Classification RBM}. In addition, their classification robustness against different types of noises have been evaluated for the MNIST images.

\subsection{Unsupervised learning}

RBMs are popular for their unsupervised learning capability and high performance as feature extractors and density estimators \cite{bengio2009learning}. Therefore, the models are trained to approximate the underlying probability distribution of handwritten digits and fashion product images, following the general pipeline described in Section \ref{sec:Pipeline}. Even though both datasets we use provide class labels, they can be considered as unlabelled and thus we train our models for generative purposes using the update rules shown in Eq. (\ref{Param Gradients}-\ref{Hidden gradient}). 

\subsubsection{Density estimation}\label{sec:LL}

\begin{figure}[t]
    \centering
    \includegraphics[width=\textwidth]{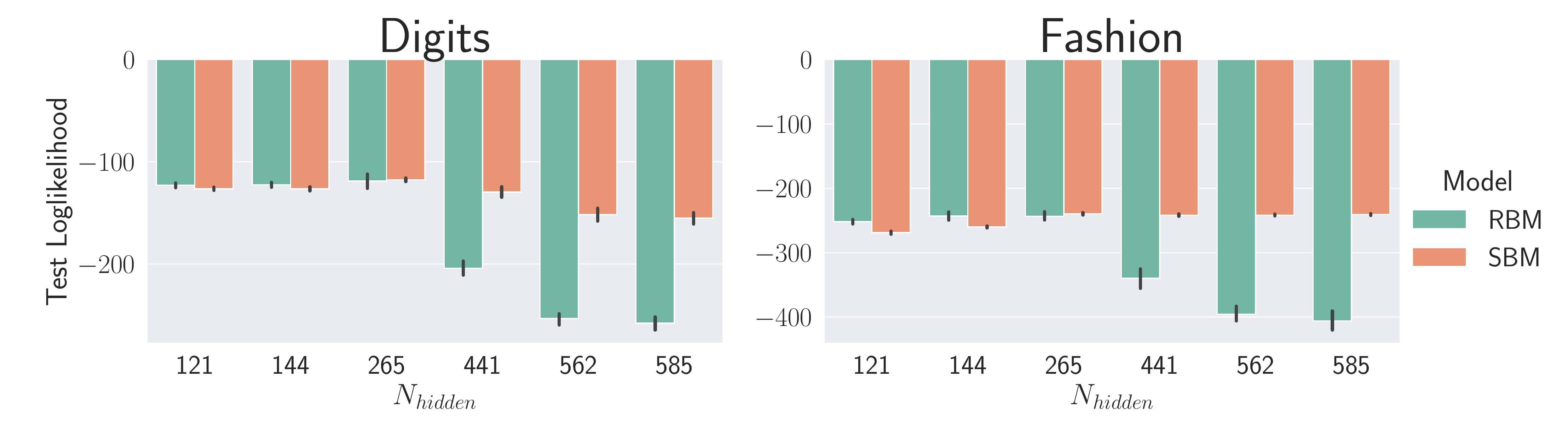}
    \caption{\textit{Loglikelihood} of the test set of images. Models were trained to estimate the probability distribution of the handwritten digits (MNIST) and fashion product images (FashionMNIST). After training, in order to assess models with the corresponding test set, parameters for which the LL of the validation set was maximum were chosen.}
    \label{Test LL}
\end{figure}

In this section, after training all models, the results of each image domain are displayed in Figure \ref{Test LL}. Comparing each \modelacronym with its respective RBM one can observe two different outcomes.

On one hand, \modelacronymp s outperform their twin RBMs on both image domains when they have more than 265 hidden units. This verifies our initial hypothesis that structured hidden-visible connections can help to enhance the performance of RBMs. On the other hand, \modelacronymp$_{121}$ and \modelacronymp$_{144}$ under-perform their twin RBM. Though this is more noticeable for images consisting of digits, in contrast to fashion products, where the LL values are more similar.

A key factor to understand these results is the neighbourhood structure of \modelacronymp s. Both under-performing models share the feature of having the fewer dense connections between hidden and visible units. A clear improvement is achieved by combining both structures, as can be seen for \modelacronymp$_{265}$, which is, in fact, the model acquired by combining both aforementioned structures. Hence, this phenomenon supports the idea of combining structures to enhance model performance.

Nevertheless, we have observed that there is a big deterioration in the test LL as RBMs increase their number of hidden neurons above 265, which is something unexpected, since their representability power should increase. In order to shed light on this matter, we analyse the evolution of the LL over the validation set during training.

\begin{figure}[t]
    \centering
    \includegraphics[width=\textwidth]{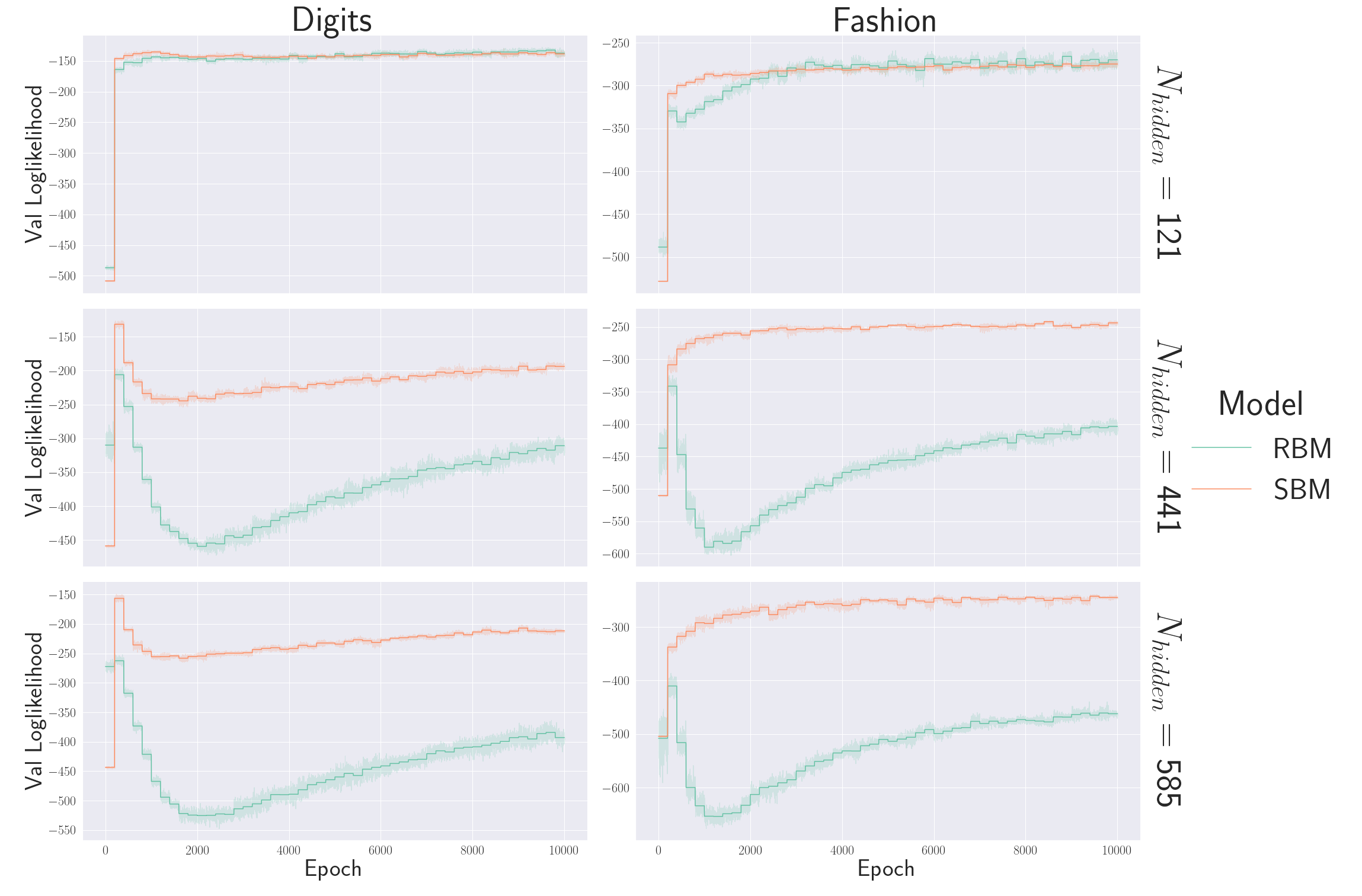}
    \caption{\textit{Loglikelihood} of the validation set during training. Y axis shows its mean value and the confidence interval over the 10 runs of the experiments with different seeds. \textit{Loglikelihood} is computed every 200 epochs in order to save time, for its computation is very time-consuming. Hence the stepped shape of the lines. }
    \label{fig:training process}
\end{figure}

Figure \ref{fig:training process} depicts the behaviour of the LL of the models over the validation set during training for $N_{hidden}\in \{121,441,585\}$. One may notice that models with more than 265 hidden units suffer from overfitting, especially the vanilla RBMs. A similar behaviour was observed in \cite{mocanu2018scalable} as well, where the proposed model was compared to RBMs. Therefore, RBMs experiencing a greater overfitting effect justifies \modelacronymp s being more robust against such an aftermath. 

Recall that, for each model, the parameters that obtained the maximum validation LL were chosen after training, in order to be evaluated with the test set. If the set of parameters achieved when training finalises was chosen, RBMs would attain even worse results, as can be observed by comparing each model validation LL values at the start of the training process with the last values. 

\subsubsection{Denoising MNIST}

\begin{figure}[ht]
    \centering
    \includegraphics[width=\textwidth]{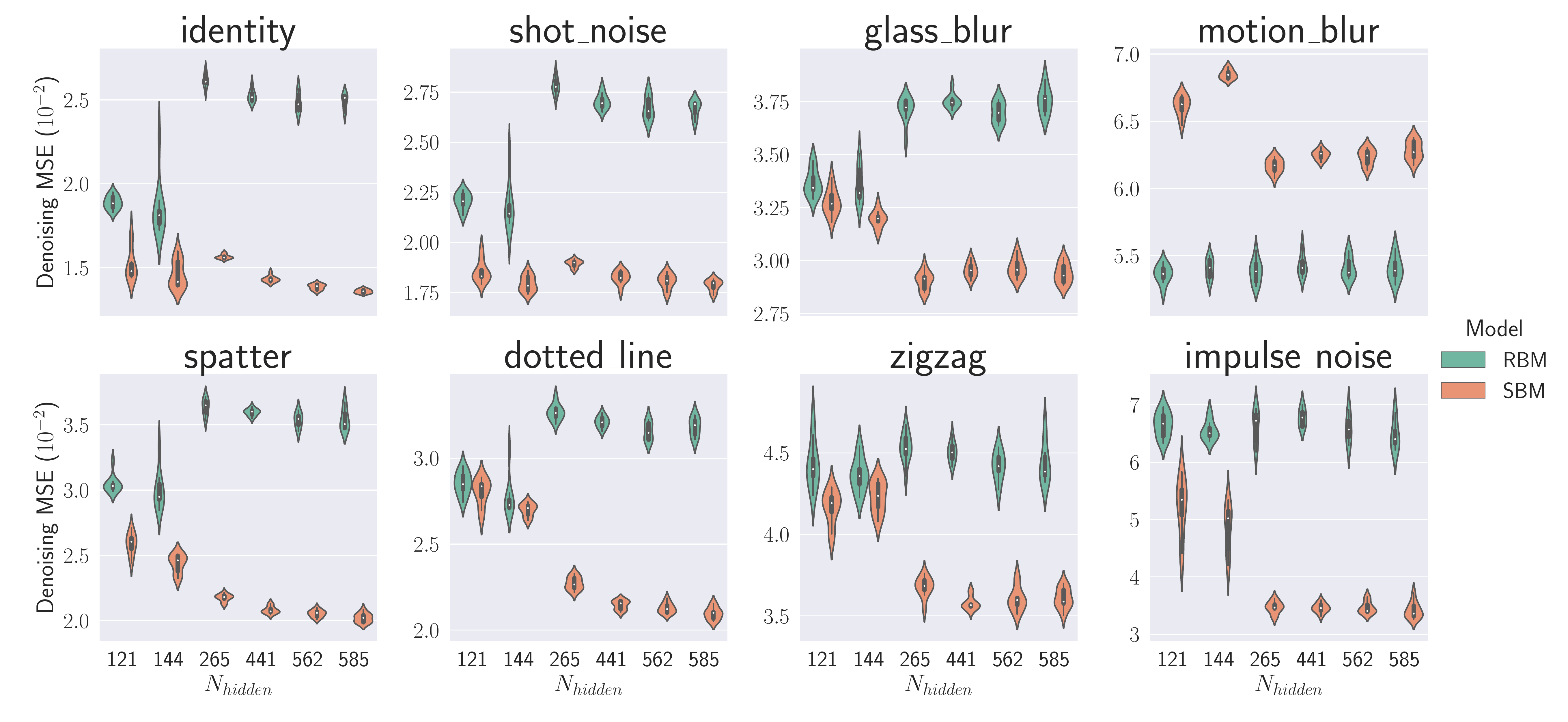}
    \caption{Denoising mean squared error (MSE) after denoising corrupted MNIST test images. The corrupted image was set as initial state of each RBM/\modelacronymp, and after a single Gibbs sampling step the obtained image is considered the reconstructed image. The \textit{identity} label refers to the uncorrupted MNIST images.}
    \label{fig:Denoising MNIST}
\end{figure}

As stated  in Section \ref{sec:evaluate RBM}, one of the Machine Learning tasks these models can handle is image denoising, consequently, we assess our models performance on this task as well. The mean denoising performance is evaluated with the aforementioned MNIST-C corrupted test images. In this manner, the performance of the model on a different unsupervised learning task can be assessed. Figure \ref{fig:Denoising MNIST} shows the results obtained with each model and noise type as violin plots \cite{Hintze_and_Nelson:1998}. It shows the distribution of the denoising MSE across several levels such that those distributions can be compared. Unlike a box plot, in which all of the plot components correspond to actual data points, the violin plot features a kernel density estimation of the results distribution. 

Figure \ref{fig:Denoising MNIST} shows that \modelacronymp s attain very similar MSE errors when denoising digit images with different noise types. There are, however, two noise types for which results are distinct.

For all noise types, \modelacronymp s obtain better denoising errors compared to their twin RBMs, except for the \textit{motion\_blur} noise. Nonetheless, \modelacronymp$_{121}$ and \modelacronymp$_{144}$ obtain the worst denoising error among \modelacronymp s. This may happen due to their specific neighbourhood structure, as discussed in Section \ref{sec:LL}. It is remarkable that \modelacronym with more than 441 hidden units obtain better results compared to RBM$_{121}$, since these \modelacronymp s have the number of trainable parameters closest to the number of parameters RBM$_{121}$ has, as can be seen in Table \ref{tab:Model params table}.

\subsubsection{Elapsed time}

It is worth recalling that \modelacronymp s used during our experiments have only between 6\% and 10\% the number of trainable parameters compared to RBMs. Thus, as authors proposed in \cite{mocanu2018scalable}, in order to speed up training time, we have implemented a faster, method where only the gradients of the non-zero weights are computed when \modelacronymp s are trained. 

\begin{figure}[t]
    \centering
    \includegraphics[width=\textwidth]{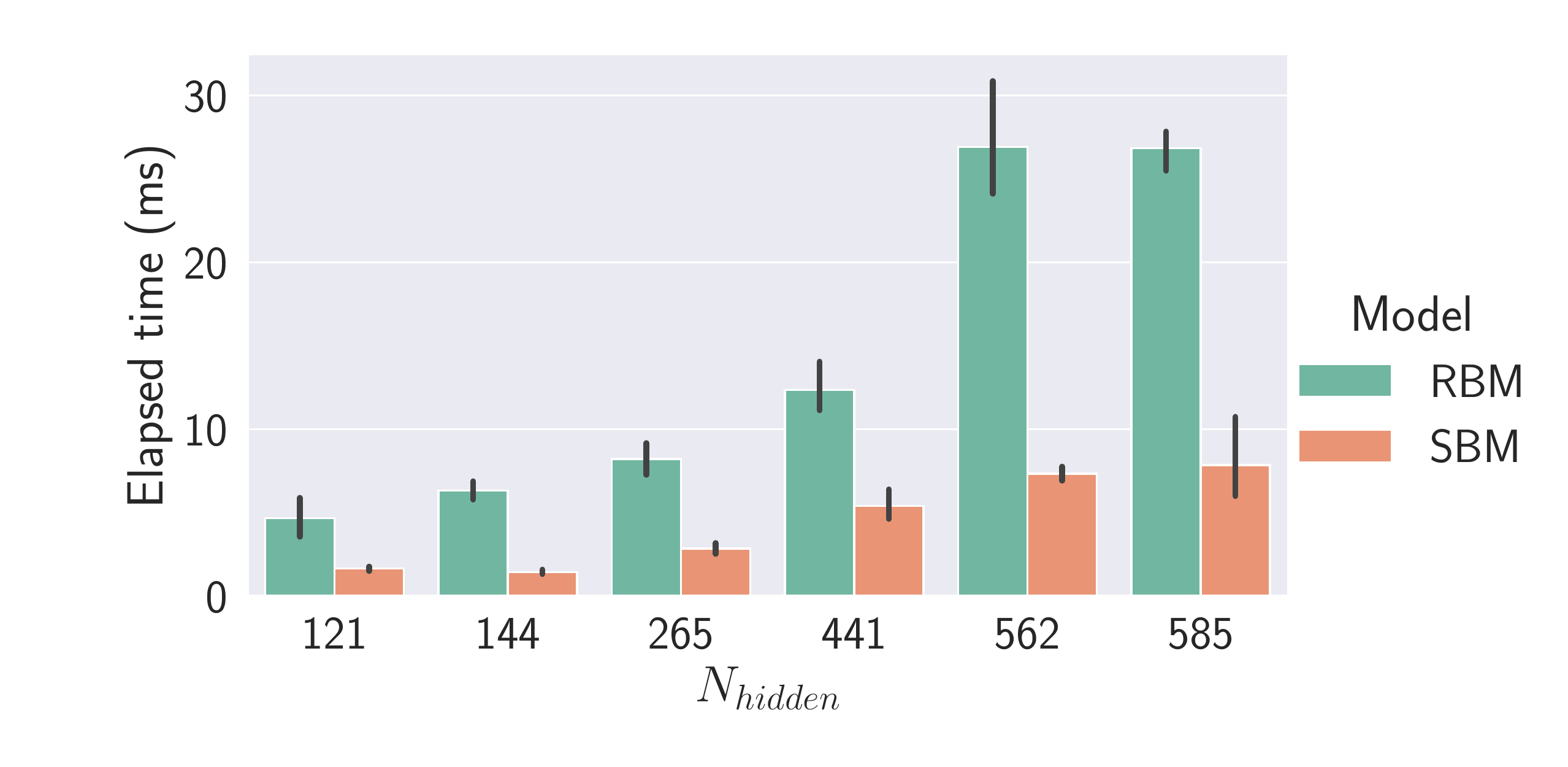}
    \caption{Mean elapsed time when computing the weight gradient for different number of hidden units, comparing an RBM with the optimised SBM update. It can be observed how, for \modelacronymp s, gradients are computed approximately three times faster compared to RBMs in every configuration.}
    \label{Elapsed time}
\end{figure}

Figure \ref{Elapsed time} presents results of the mean elapsed time over all training epochs when computing the gradients on Eq. (\ref{W gradient}). The vertical black lines over each bar depict the standard deviation of those mean values over the 10 repetitions of the experiments performed in the previous section.

As expected, gradients are computed much faster with \modelacronymp s. It takes approximately three times more time to compute gradients with RBMs, which implies slower updates of parameters, and thus, a slower training compared to our proposed model.

\subsection{Classification performance}

\begin{figure}[t]
    \centering
    \includegraphics[width=\textwidth]{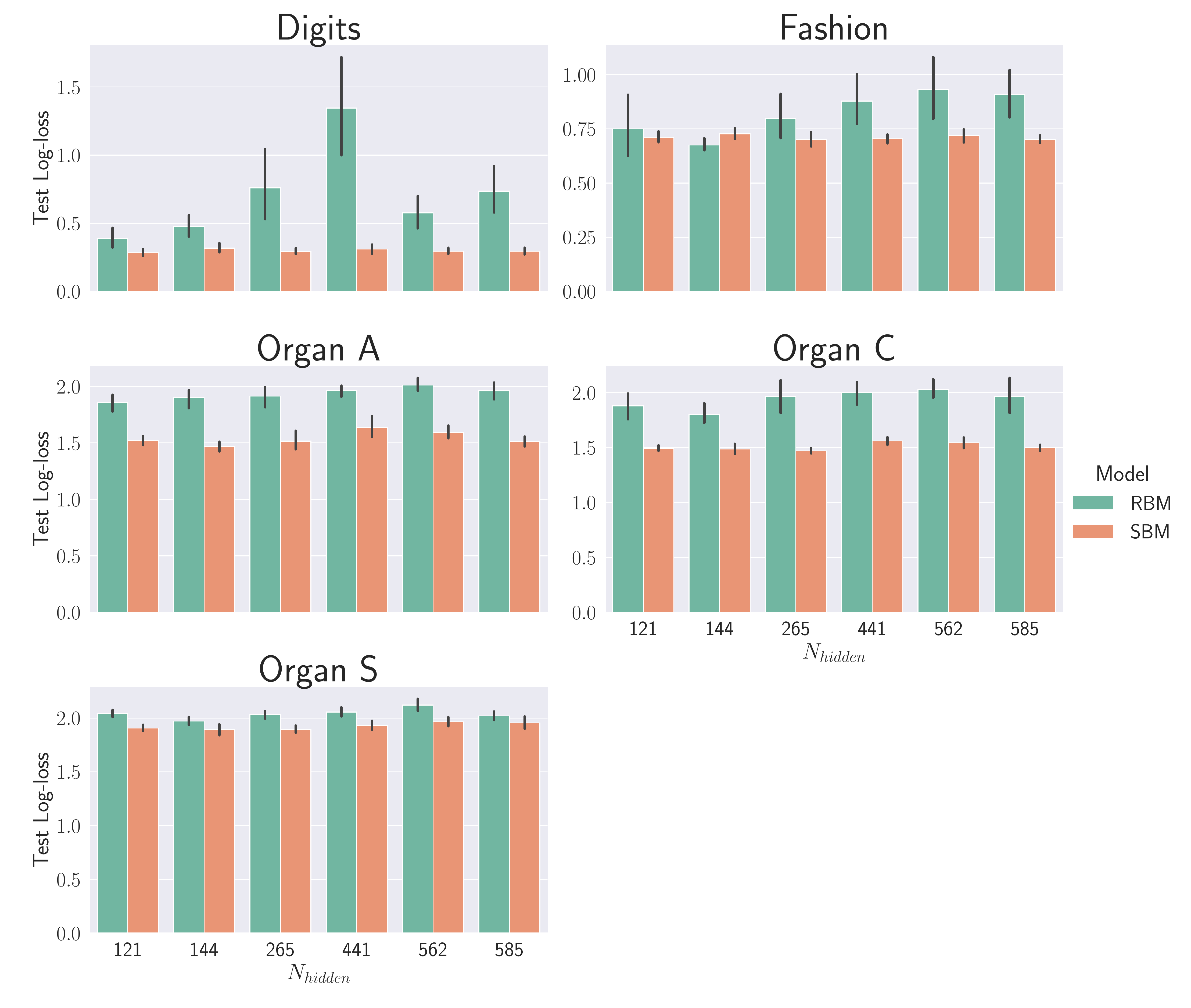}
    \caption{Log-loss values of the respective test set for each dataset. For medical images, the balanced Log-loss was computed, since datasets present unbalanced classes.}
    \label{fig:Logloss}
\end{figure}

\begin{figure}[t]
    \centering
    \includegraphics[width=\textwidth]{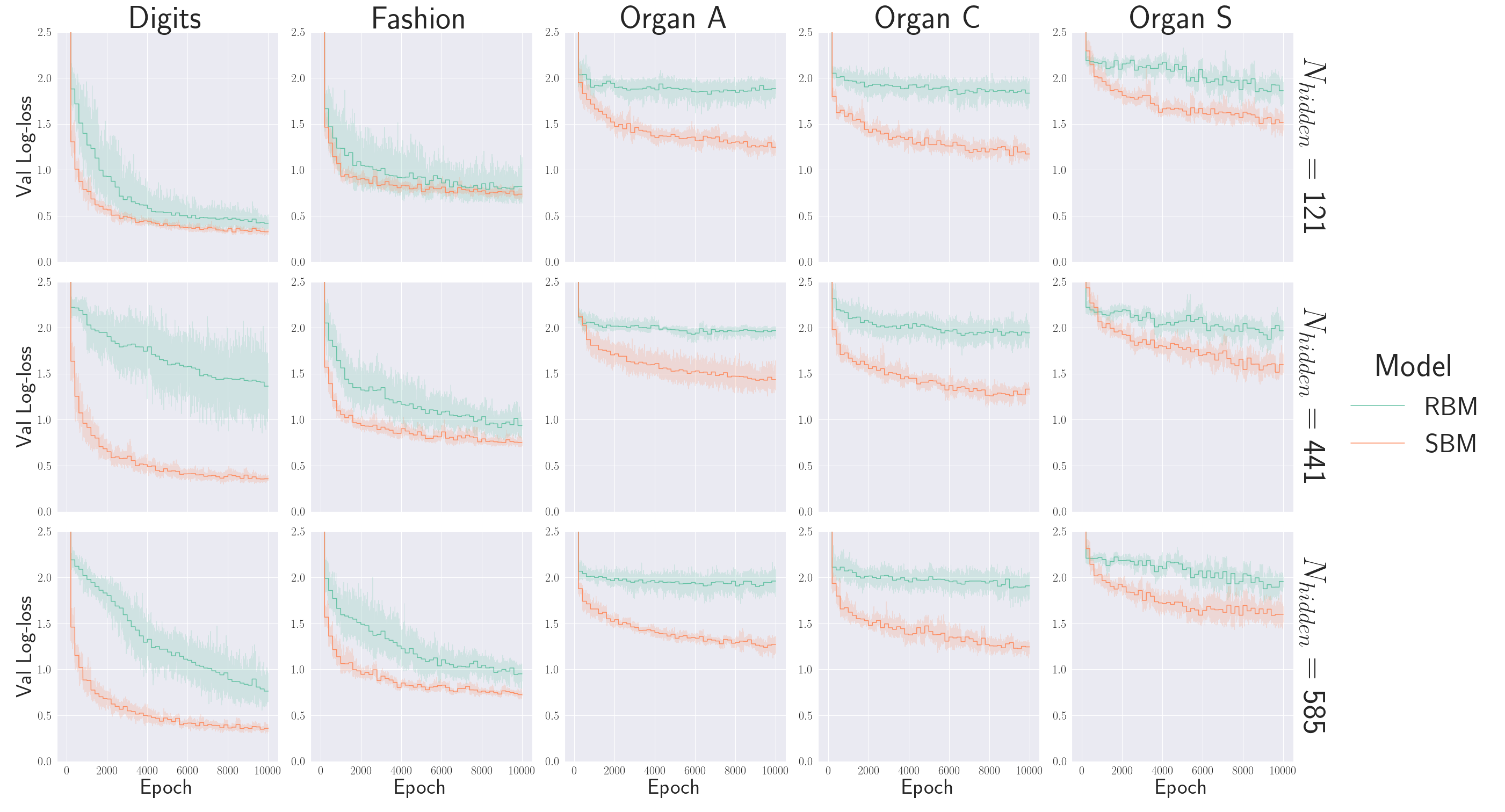}
    \caption{Log-loss of the validation set during training. Y axis shows its mean value  and the confidence interval over the 10 runs of the experiments with different seeds. Log-loss was computed every 200 epochs in order to save time. For medical images, the balanced version of the Log-loss was computed since classes are unbalanced.}
    \label{fig:Logloss during training}
\end{figure}

Classifying images is one of the most widespread and important problems nowadays, hence we further explore the performance of our models by addressing the image classification problem. As stated in Section \ref{Classification RBM}, RBMs can be used as discriminative models. Thus, we can compute the probability of a given image being of a certain class. We have trained our models to classify handwritten digits, fashion products and three organ based medical images from MedMNIST.

Figure \ref{fig:Logloss} presents results of the Log-loss values of the test set after training models with the discriminative update rules in Eq. (\ref{Classification gradients}). These experiments highlight the fact that \modelacronymp s outperform RBMs in plenty of image domains and problems, thus making them more appropriate to use in classification tasks as well. Fashion product images being the only exception where, for \modelacronymp s with less than 265 hidden units, RBMs obtain smaller values of Log-loss values. 

Moreover, we highlight the fact that \modelacronym with more than 441 hidden units, having a similar number of trainable parameters compared to RBM$_{121}$, outperform RBM$_{121}$ in all experiments.

It is remarkable that RBM$_{441}$ spikes in Log-loss value when classifying digits compared to every other model, and thus has the worst performance in this specific scenario. Similar to the previous section, Figure \ref{fig:Logloss during training} shows the Log-loss of the validation set during training for $N_{hidden}\in \{121,441,585\}$, in order to illustrate what could have happened to RBM$_{441}$.

As can be seen in Figure \ref{fig:Logloss during training}, RBM$_{441}$'s Log-loss over the validation decreases slowly when trained with digit images after some epochs, which explains its bad performance. For every other model and image domain, we observe that \modelacronymp s always have smaller values of Log-loss compared to their twin RBMs. With the exception of fashion product images, where RBMs and \modelacronymp s obtain comparable Log-loss values during training, thus explaining their similar results when assessed with the test set afterwards. Furthermore, the confidence interval illustrates how RBMs have a more unstable training compared to \modelacronymp s.  

\subsubsection{Classification robustness against noise}

We have evaluated how \modelacronym performs compared to their twin RBM for different datasets, and we would also like to assess the robustness of their classification performance against noisy data. This is an important additional analysis to the classification problem, because it showcases how well a model behaves against potential real data which commonly involves some kind of noise. The MNIST-C dataset was in fact devised to assess the robustness of classifiers against different noises when classifying MNIST images. Hence, we evaluate the classification performance of the models trained with MNIST, computing the Log-loss obtained with the corrupted images used in the denoising experiments.  

\begin{figure}[t]
    \centering
    \includegraphics[width=\textwidth]{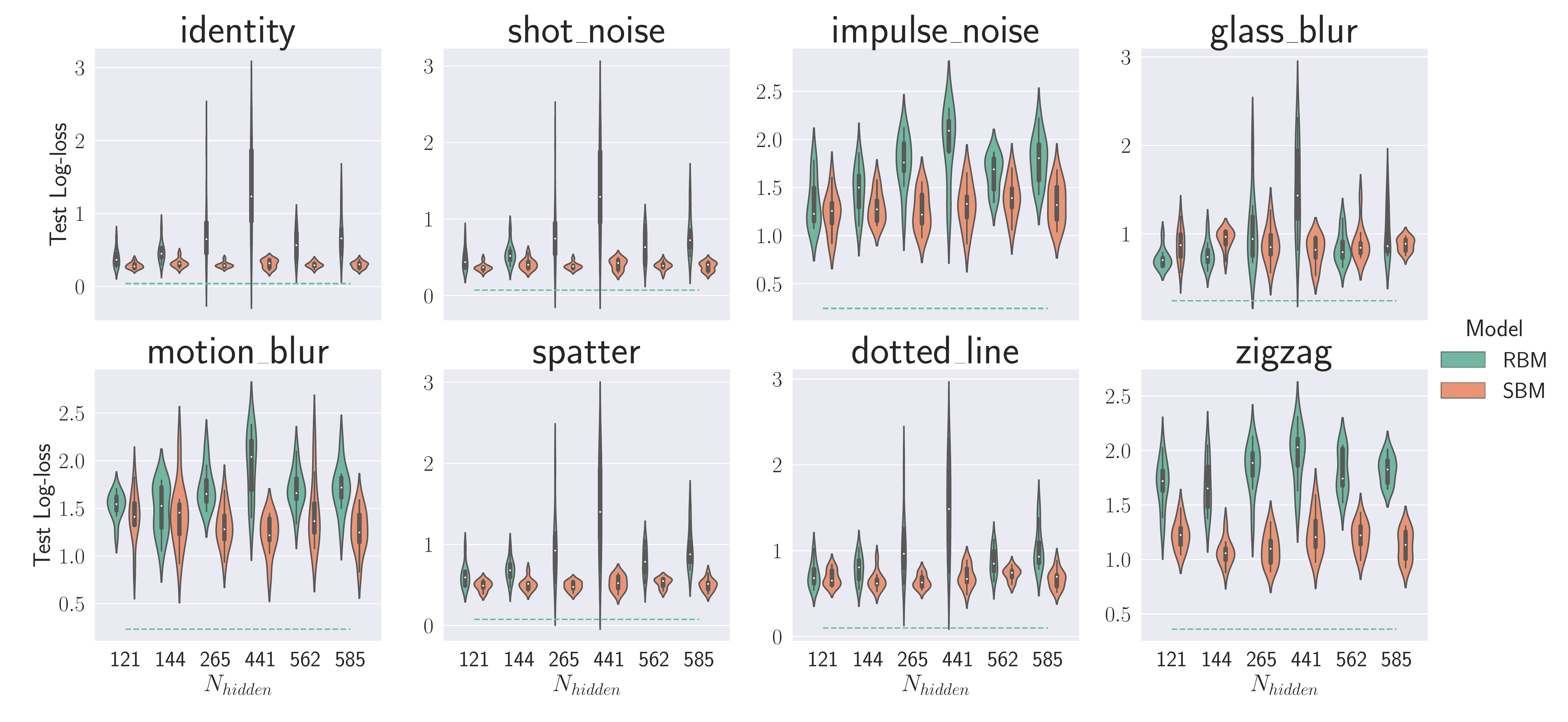}
    \caption{Log-loss of the MNIST-C test images for each type of corruption, using the previously trained models to classify the original MNIST images. A CNN was trained to classify MNIST images, this way the Log-loss value that the CNN obtains for each corruption type was compared to our results. The horizontal dashed lines indicate said values.}
    \label{fig:Robustness MNIST}
\end{figure}

In Figure \ref{fig:Robustness MNIST}, the Log-loss values for each type of noise are displayed. Additionally, a CNN\footnote{The model is available at: \url{https://github.com/AmritK10/MNIST-CNN}.} was trained to compare how it performed against these corruptions, so that we were able to contrast how other models behave against these noises. The Log-loss value of corrupted images computed with the CNN are added in Figure \ref{fig:Robustness MNIST} as a horizontal dashed line for reference.

Yet again, \modelacronymp s obtain very similar or even better results compared to their twin RBMs. It can be observed that RBM$_{441}$ obtains the highest (worst) Log-loss values, this could be expected since we have already observed its poor results in Figure \ref{fig:Logloss}, when evaluating its performance with handwritten digit images. 

Comparing our model's results with those obtained by the CNN, we can say that, with noises \textit{shot\_noise}, \textit{glass\_blur}, \textit{spatter} and \textit{dotted\_line}, our models performance and that of CNN do not differ as much as they do with other noise types. The classification performance is also similar when classifying the original images (\textit{identity}). For every other noise type, the CNN achieves much better Log-loss values. However, it is remarkable the fact that the CNN has $3\cdot 10^6$ trainable parameters, which is 10 times bigger than the largest RBM used in our experiments.

\section{Conclusions and Future Work}\label{Conclusion}

In this paper \modelacronymp s have been introduced. They require less trainable parameters than RBMs. We have proposed a method to create sparse connections using information of the problem to be solved. As an area of application, we have trained our models to address standard Machine Learning problems involving images, and thus the structure of sparse connections is inspired by prior knowledge about images. Experiments show evidence that our models are faster to train and achieve even better results compared to RBMs.

Experiments on unsupervised learning tasks show that, for \modelacronymp s with dense enough connections, results are improved compared to their twin RBMs. Additionally, combining structures has proven to be effective in order to attain better results. Moreover, due to the sparse nature of our model's parameters, having between 94\% and 90\% less trainable parameters compared to their twin RBMs, training \modelacronymp s is much faster compared to training vanilla RBMs. In fact, it has been empirically demonstrated that \modelacronymp s are more robust against the overfitting effect, even when having a large number of hidden units, in contrast to RBMs. What is more, it was observed that comparing some \modelacronymp s and an RBM with a similar number of trainable parameters, the \modelacronymp s obtained better results than the RBM, which supports the use of our models. The classification problem has been investigated using 5 different image domains, where \modelacronymp s presented better achievements as well. The classification robustness against noisy data of the proposed model has been contrasted with RBMs, where again, \modelacronymp s present comparable results to RBMs performance. 

For future work, alternatives to CD can be assessed when training, for instance increasing the number of Gibbs sampling steps or using other methods such as Ratio matching \cite{hyvarinen2007some}. In our experiments, the data consisted of images, and therefore the structure of sparse weights was proposed for grey-scale images. However, structures for more complex data can be proposed, such as RGB images or 3D data. Moreover, removing the man-in-the-loop is one of our main concerns, hence methods that automatically learn a neighbourhood structure from the data will be studied.

\section*{Acknowledgements}
The authors would like to thank Dr. Decebal Constantin Mocanu for facilitating the code for optimising gradient computations for sparse RBMs in Cython, in which we have based our code. Arkaitz Bidaurrazaga and Aritz Pérez have been supported by the Basque Government through the BERC 2022-2025 program and by the Ministry of Science, Innovation and Universities: BCAM Severo Ochoa accreditation SEV-2017-0718. Roberto Santana's work is supported by the Basque Government 
(KK2020/00049 project through ELKARTEK program), and by the Spanish 
Ministry of Science, Innovation and Universities (project 
PID2019-104966GB-I00).

\bibliographystyle{ieeetr}
\bibliography{Main.bbl}

\begin{thebibliography}{10}

\bibitem{gehler2006rate}
P.~V. Gehler, A.~D. Holub, and M.~Welling, ``{The Rate Adapting Poisson Model
  for Information Retrieval and Object Recognition},'' in {\em Proceedings of
  the 23rd {I}nternational {C}onference on {M}achine {L}earning}, pp.~337--344,
  2006.

\bibitem{hinton2007recognize}
G.~E. Hinton, ``To recognize shapes, first learn to generate images,'' {\em
  {Progress in Brain Research}}, vol.~165, pp.~535--547, 2007.

\bibitem{hopfield1982neural}
J.~J. Hopfield, ``Neural networks and physical systems with emergent collective
  computational abilities,'' {\em Proceedings of the National Academy of
  Sciences}, vol.~79, no.~8, pp.~2554--2558, 1982.

\bibitem{hinton1983optimal}
G.~E. Hinton and T.~J. Sejnowski, ``Optimal perceptual inference,'' in {\em
  Proceedings of the IEEE Conference on Computer Vision and Pattern
  Recognition}, vol.~448, pp.~448--453, 1983.

\bibitem{smolensky1986information}
P.~Smolensky, ``Information processing in dynamical systems: Foundations of
  harmony theory,'' tech. rep., Colorado Univ. at Boulder Dept. of Computer
  Science, 1986.

\bibitem{welling2004exponential}
M.~Welling, M.~Rosen-Zvi, and G.~E. Hinton, ``Exponential family harmoniums
  with an application to information retrieval,'' {\em Advances in {N}eural
  {I}nformation {P}rocessing {S}ystems}, vol.~17, 2004.

\bibitem{courville2011spike}
A.~Courville, J.~Bergstra, and Y.~Bengio, ``{A spike and slab restricted
  Boltzmann machine},'' in {\em {Proceedings of the Fourteenth International
  Conference on Artificial Intelligence and Statistics}}, pp.~233--241, JMLR
  Workshop and Conference Proceedings, 2011.

\bibitem{montufar2016restricted}
G.~Mont{\'u}far, ``Restricted {B}oltzmann machines: {I}ntroduction and
  review,'' in {\em Information Geometry and Its Applications IV}, pp.~75--115,
  Springer, 2016.

\bibitem{upadhya2019overview}
V.~Upadhya and P.~Sastry, ``An overview of restricted {B}oltzmann machines,''
  {\em Journal of the Indian Institute of Science}, pp.~1--12, 2019.

\bibitem{hinton2002training}
G.~E. Hinton, ``Training products of experts by minimizing contrastive
  divergence,'' {\em Neural Computation}, vol.~14, no.~8, pp.~1771--1800, 2002.

\bibitem{geman1984stochastic}
S.~Geman and D.~Geman, ``Stochastic relaxation, {G}ibbs distributions, and the
  {B}ayesian restoration of images,'' {\em IEEE Transactions on Pattern
  Analysis and Machine Intelligence}, no.~6, pp.~721--741, 1984.

\bibitem{carreira2005contrastive}
M.~A. Carreira-Perpinan and G.~Hinton, ``On contrastive divergence learning,''
  in {\em International Workshop on Artificial Intelligence and Statistics},
  pp.~33--40, PMLR, 2005.

\bibitem{loukas2019self}
O.~Loukas, ``Self-regularizing restricted {B}oltzmann machines,'' {\em arXiv
  preprint arXiv:1912.05634}, 2019.

\bibitem{salakhutdinov2008quantitative}
R.~Salakhutdinov and I.~Murray, ``On the quantitative analysis of deep belief
  networks,'' in {\em Proceedings of the 25th International Conference on
  Machine Learning}, pp.~872--879, 2008.

\bibitem{larochelle2008classification}
H.~Larochelle and Y.~Bengio, ``Classification using discriminative {R}estricted
  {B}oltzmann machines,'' in {\em Proceedings of the 25th International
  Conference on Machine Learning}, pp.~536--543, 2008.

\bibitem{bi2019early}
X.~Bi and H.~Wang, ``Early {A}lzheimer’s disease diagnosis based on {EEG}
  spectral images using deep learning,'' {\em Neural Networks}, vol.~114,
  pp.~119--135, 2019.

\bibitem{he2013facial}
S.~He, S.~Wang, W.~Lan, H.~Fu, and Q.~Ji, ``Facial expression recognition using
  deep {B}oltzmann machine from thermal infrared images,'' in {\em 2013 Humaine
  Association Conference on Affective Computing and Intelligent Interaction},
  pp.~239--244, IEEE, 2013.

\bibitem{pilati2020simulating}
S.~Pilati and P.~Pieri, ``Simulating disordered quantum {I}sing chains via
  dense and sparse {R}estricted {B}oltzmann machines,'' {\em Physical Review
  E}, vol.~101, no.~6, p.~063308, 2020.

\bibitem{sehayek2019learnability}
D.~Sehayek, A.~Golubeva, M.~S. Albergo, B.~Kulchytskyy, G.~Torlai, and R.~G.
  Melko, ``Learnability scaling of quantum states: Restricted {B}oltzmann
  machines,'' {\em Physical Review B}, vol.~100, no.~19, p.~195125, 2019.

\bibitem{mittelman2014structured}
R.~Mittelman, B.~Kuipers, S.~Savarese, and H.~Lee, ``Structured recurrent
  temporal restricted {B}oltzmann machines,'' in {\em International Conference
  on Machine Learning}, pp.~1647--1655, 2014.

\bibitem{ji2014enhancing}
N.~Ji, J.~Zhang, C.~Zhang, and Q.~Yin, ``{Enhancing performance of restricted
  Boltzmann machines via log-sum regularization},'' {\em Knowledge-Based
  Systems}, vol.~63, pp.~82--96, 2014.

\bibitem{cho2012tikhonov}
K.~Cho, A.~Ilin, and T.~Raiko, ``{Tikhonov-type regularization for restricted
  Boltzmann machines},'' in {\em International Conference on Artificial Neural
  Networks}, pp.~81--88, Springer, 2012.

\bibitem{mocanu2016topological}
D.~C. Mocanu, E.~Mocanu, P.~H. Nguyen, M.~Gibescu, and A.~Liotta, ``A
  topological insight into {R}estricted {B}oltzmann machines,'' {\em Machine
  Learning}, vol.~104, no.~2, pp.~243--270, 2016.

\bibitem{mocanu2018scalable}
D.~C. Mocanu, E.~Mocanu, P.~Stone, P.~H. Nguyen, M.~Gibescu, and A.~Liotta,
  ``Scalable training of artificial neural networks with adaptive sparse
  connectivity inspired by network science,'' {\em Nature {C}ommunications},
  vol.~9, no.~1, pp.~1--12, 2018.

\bibitem{kunsch1995hidden}
H.~Kunsch, S.~Geman, and A.~Kehagias, ``Hidden {M}arkov random fields,'' {\em
  The Annals of Applied Probability}, vol.~5, no.~3, pp.~577--602, 1995.

\bibitem{frankle2019lottery}
J.~Frankle and M.~Carbin, ``The lottery ticket hypothesis: Finding sparse,
  trainable neural networks,'' {\em arXiv preprint arXiv:1803.03635}, 2018.

\bibitem{lee2019snip}
N.~Lee, T.~Ajanthan, and P.~H. Torr, ``Snip: Single-shot network pruning based
  on connection sensitivity,'' {\em arXiv preprint arXiv:1810.02340}, 2018.

\bibitem{bengio2000taking}
S.~Bengio and Y.~Bengio, ``Taking on the curse of dimensionality in joint
  distributions using neural networks,'' {\em IEEE Transactions on Neural
  Networks}, vol.~11, no.~3, pp.~550--557, 2000.

\bibitem{jain2020locally}
A.~Jain, P.~Abbeel, and D.~Pathak, ``Locally masked convolution for
  autoregressive models,'' in {\em Conference on Uncertainty in Artificial
  Intelligence}, pp.~1358--1367, PMLR, 2020.

\bibitem{mayraz2000recognizing}
G.~Mayraz and G.~E. Hinton, ``Recognizing hand-written digits using
  hierarchical products of experts,'' {\em Advances in Neural Information
  Processing Systems}, vol.~13, 2000.

\bibitem{mu2019mnist}
N.~Mu and J.~Gilmer, ``{MNIST}-{C}: {A} robustness benchmark for computer
  vision,'' {\em arXiv preprint arXiv:1906.02337}, 2019.

\bibitem{lecun1998gradient}
Y.~LeCun, L.~Bottou, Y.~Bengio, and P.~Haffner, ``Gradient-based learning
  applied to document recognition,'' {\em Proceedings of the IEEE}, vol.~86,
  no.~11, pp.~2278--2324, 1998.

\bibitem{DBLP:journals/corr/abs-1708-07747}
H.~Xiao, K.~Rasul, and R.~Vollgraf, ``Fashion-{MNIST}: a novel image dataset
  for benchmarking machine learning algorithms,'' {\em CoRR},
  vol.~abs/1708.07747, 2017.

\bibitem{medmnistv1}
J.~Yang, R.~Shi, and B.~Ni, ``Med{MNIST} {C}lassification {D}ecathlon: {A}
  {L}ightweight {AutoML} {B}enchmark for {M}edical {I}mage {A}nalysis,'' in
  {\em IEEE 18th International Symposium on Biomedical Imaging (ISBI)},
  pp.~191--195, 2021.

\bibitem{tensorflow2015}
M.~Abadi, A.~Agarwal, P.~Barham, E.~Brevdo, Z.~Chen, C.~Citro, G.~S. Corrado,
  A.~Davis, J.~Dean, M.~Devin, S.~Ghemawat, I.~Goodfellow, A.~Harp, G.~Irving,
  M.~Isard, Y.~Jia, R.~Jozefowicz, L.~Kaiser, M.~Kudlur, J.~Levenberg,
  D.~Man\'{e}, R.~Monga, S.~Moore, D.~Murray, C.~Olah, M.~Schuster, J.~Shlens,
  B.~Steiner, I.~Sutskever, K.~Talwar, P.~Tucker, V.~Vanhoucke, V.~Vasudevan,
  F.~Vi\'{e}gas, O.~Vinyals, P.~Warden, M.~Wattenberg, M.~Wicke, Y.~Yu, and
  X.~Zheng, ``{TensorFlow}: Large-scale machine learning on heterogeneous
  systems,'' 2015.
\newblock Software available from tensorflow.org.

\bibitem{qian1999momentum}
N.~Qian, ``On the momentum term in gradient descent learning algorithms,'' {\em
  Neural networks}, vol.~12, no.~1, pp.~145--151, 1999.

\bibitem{glorot2010understanding}
X.~Glorot and Y.~Bengio, ``Understanding the difficulty of training deep
  feedforward neural networks,'' in {\em Proceedings of the Thirteenth
  International Conference on Artificial Intelligence and Statistics},
  pp.~249--256, JMLR Workshop and Conference Proceedings, 2010.

\bibitem{hinton2012practical}
G.~E. Hinton, ``A practical guide to training {R}estricted {B}oltzmann
  machines,'' in {\em Neural Networks: Tricks of the Trade}, pp.~599--619,
  Springer, 2012.

\bibitem{bengio2009learning}
Y.~Bengio {\em et~al.}, ``{Learning deep architectures for AI},'' {\em
  Foundations and trends{\textregistered} in Machine Learning}, vol.~2, no.~1,
  pp.~1--127, 2009.

\bibitem{Hintze_and_Nelson:1998}
J.~L. Hintze and R.~D. Nelson, ``Violin plots: a box plot-density trace
  synergism,'' {\em The American Statistician}, vol.~52, no.~2, pp.~181--184,
  1998.

\bibitem{hyvarinen2007some}
A.~Hyv{\"a}rinen, ``Some extensions of {S}core {M}atching,'' {\em Computational
  {S}tatistics \& {D}ata {A}nalysis}, vol.~51, no.~5, pp.~2499--2512, 2007.

\end{thebibliography}
\end{document}